%% file: main.tex
\documentclass{article}
\usepackage{arxiv}

\usepackage[utf8]{inputenc} 
\usepackage[T1]{fontenc}    
\usepackage{natbib}
\usepackage{graphicx}
\usepackage{amsmath}
\usepackage{hyperref}

\usepackage{tabularx,booktabs,multirow,array}
\AtBeginDocument{%
  }

\usepackage{listings}
\usepackage{algorithm}
\usepackage{algorithmicx}
\usepackage{algpseudocode}

\begin{document}

\title{Personalized Fall Detection by Balancing Data with Selective Feedback Using Contrastive Learning}

\author{
    Awatif Yasmin \\
    Texas State University \\
  \texttt{nuc4@txstate.edu} \\
    \And
 Tarek Mahmud \\
  Texas A\&M University - Kingsville\\
  \texttt{tarek.mahmud@tamuk.edu} \\
  \And
 Sana Alamgeer \\
  Texas State University \\ 
  \texttt{sana.alamgeer@txstate.edu} \\
  \And
 Anne H. H. Ngu \\
  Texas State University \\
  \texttt{angu@txstate.edu} \\
}



\maketitle

\begin{abstract}
Personalized fall detection models can significantly improve accuracy by adapting to individual motion patterns, yet their effectiveness is often limited by the scarcity of real-world fall data and the dominance of non-fall feedback samples. This imbalance biases the model toward routine activities and weakens its sensitivity to true fall events. To address this challenge, we propose a personalization framework that combines semi-supervised clustering with contrastive learning to identify and balance the most informative user feedback samples. The framework is evaluated under three retraining strategies, including Training from Scratch (TFS), Transfer Learning (TL), and Few-Shot Learning (FSL), to assess adaptability across learning paradigms. Real-time experiments with ten participants show that the TFS approach achieves the highest performance, with up to a 25\% improvement over the baseline, while FSL achieves the second-highest performance with a 7\% improvement, demonstrating the effectiveness of selective personalization for real-world deployment.
\end{abstract}

\keywords{Fall Detection, Personalization, Transformer}

\input{sections/intro}
\input{sections/motivation}

\input{sections/related_work}
\input{sections/method}
\input{sections/experimental_design}
\input{sections/results}
\input{sections/conclusion}
\bibliographystyle{plain}
\bibliography{Bibliography}

\end{document}

%% file: sections/intro.tex
\section{Introduction}
\label{sec:introduction}


Personalization has become a cornerstone of modern intelligent systems, playing a crucial role in enhancing user experiences and system performance across a wide range of domains, from content recommendation engines (like those used in YouTube) to adaptive smart homes \cite{keyanfar2024using, safaei2024deeplt} and autonomous vehicles \cite{karagulle2024incorporating, kolasani2024connected, 10415518}. Personalization allows systems to tailor their behavior to individual users' unique preferences, behaviors, and contexts \cite{mhalla2024domain}. This user-centric approach is particularly valuable in applications where variability in user characteristics or environments can significantly affect the system's effectiveness. By learning and adapting to individual patterns over time, personalized models can offer improved accuracy, applicability, and usability compared to generalized, one-size-fits-all solutions.

In the healthcare domain, personalization is especially critical, as physiological, behavioral, and environmental factors vary greatly from user to user \cite{li2024innovation}. A personalized fall detection system is beneficial for monitoring elderly individuals living independently, as timely and accurate detection is essential to ensure rapid assistance.  Traditional fall detection systems often rely on generalized models that fail to account for individual differences in movement patterns.
As a result, these systems will result in high false alarms or miss actual fall events, and result in low user acceptance.


To build an accurate and practical personalized fall detection model, it is essential to collect motion-based activity data from the target user. Several recent works propose personalization strategies \cite{ferrari2020personalization, mhalla2024domain, ferrari2023deep, diraco2023human} for human activity recognition and fall detection that rely on user metadata, domain adaptation, labeled retraining, and multi-sensor setups, respectively, which are not feasible due to their dependence on external data, continuous data assumptions, and complex infrastructure. Additionally, fall events are usually rare, making it difficult to gather fall-specific data during normal usage.  When a user wears a sensor-embedded device for fall detection, the majority of the data collected typically consists of Activities of Daily Living (ADLs), with little to no fall instances. Retraining the model using only this ADL data creates a highly imbalanced dataset, which biases the model toward recognizing ADLs and reduces its sensitivity to falls. 

The issue becomes more severe with each round of personalization when additional ADL samples are added without corresponding fall data, where only a few types of ADL samples dominate (i.e., walking, sitting, etc.). 

To mitigate this imbalance, we employ semi-supervised clustering with contrastive learning to selectively choose and balance the training data to train a personalized model, ensuring that the personalized models are trained on representative and evenly distributed data. 
This approach also helps maintain a higher recall compared to the one-size-fits-all fall detection model. We evaluated the effectiveness of our approach using the SmartFallMM dataset \cite{smartfallMM}. We trained the initial fall detection model using Transformer ($M_O$), which serves as the baseline for all participants.
Subsequently, each user's feedback data was gathered for 18 hours while the user wore the smartwatch on his/her left wrist and interacted with the system in the home environment. During this time, the initial model $M_O$ was running in the background. When a fall was predicted by the system, users confirm whether it was a true positive or a false alarm, enabling the system to label and store the feedback data. This data was then used to retrain the $M_O$ for each individual, resulting in personalized models ($M_{Pi}$, where $i \in \{1, 2, \dots, N\}$ and $N$ is the total number of participants).  
Our experimental results showed that all the personalized models $M_{Pi}$ trained with selectively chosen ADL data achieved better precision and recall than the initial common model $M_O$ as well as the models trained with combined feedback data from all users.



The primary contributions of this work are summarized as follows:
\begin{itemize}
    \item Balanced Data for Personalization:  
    We use a semi-supervised contrastive learning to carefully select ADL data for re-training. This approach not only addresses the overall class imbalance between fall and non-fall activities but also ensures diversity within the non-fall class itself. By avoiding over-representation of any single activity (e.g., walking or sitting), we create a more balanced and representative training set, which improves the quality and generalizability of the personalized model, $M_{Pi}$.

    \item Improved Recall in Personalized Models:  
    Our method shows a significant improvement in recall compared to the model trained with combined feedback data from all users. This means the personalized models trained with our selective feedback data, $M_{Pi}$, are better at detecting fall events without increasing false positives, which is important for real-world scenarios.

    \item Practical Personalization Process:  
    The proposed framework simplifies and scales the process of personalization by embedding semi-supervised learning in the pipeline.  By automatically selecting training samples using semi-supervised clustering and similarity metrics, we eliminate the need for manual labeling of ADL feedback data for re-training. 
\end{itemize}

%% file: sections/motivation.tex
\section{Background} \label{sec:background}

\begin{figure}[t]   
    \centering
    \includegraphics[width=.60\textwidth]{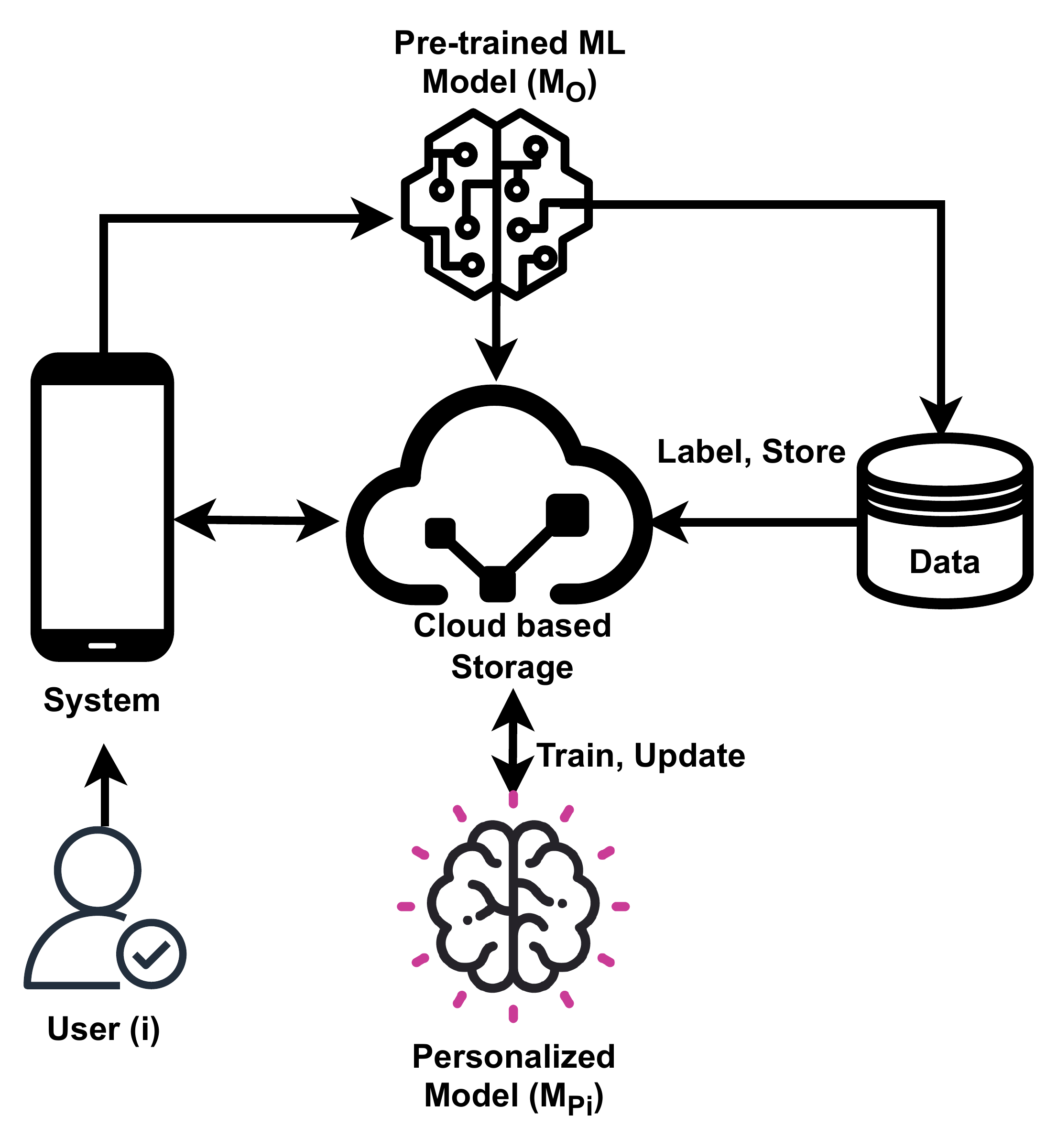}
     \caption{Personalization pipeline}
     \label{fig: pipeline}
 \end{figure}  
 
Personalization is a strategy to optimize machine learning models for individual users or specific user groups, particularly in scenarios where generalized, one-size-fits-all models fail to achieve high accuracy \cite{ferrari2020personalization}. This approach becomes essential in domains involving biometric and behavioral data, such as human activity recognition, voice-prints, gait cycles, and hand movement patterns, including orientation and grasp, where the inherent variability across individuals makes it difficult for models to generalize effectively.


In such systems, a typical pipeline, as shown in figure~\ref{fig: pipeline}, begins with the user $i$ interacting with a device or environment equipped with sensors, such as smartphones, smartwatches, cameras, or ambient embedded systems within smart homes. During initial usage, the system collects demographic metadata and a limited set of real-time sensor data to serve as a calibration phase. Pre-trained fall detection model ($M_O$) makes early predictions based on this data. Concurrently, the system continuously stores the sensed time series data and predicted labels on a cloud-based server. Once a sufficient volume of data has been collected from a particular user or group based on a defined metric, a personalized model ($M_{Pi}$ corresponding to the $i$-th user) is trained.
This updated model is then stored in a cloud infrastructure and deployed back to the device when the user activates the App next, offering improved accuracy tailored to that user’s specific behavioral patterns. As more user data is collected, the repeated process of retraining helps the model become more precise with a reduced  number of false positives without sacrificing the recall.

\begin{figure}[t!]
\centering
    \includegraphics[width=.7\columnwidth]{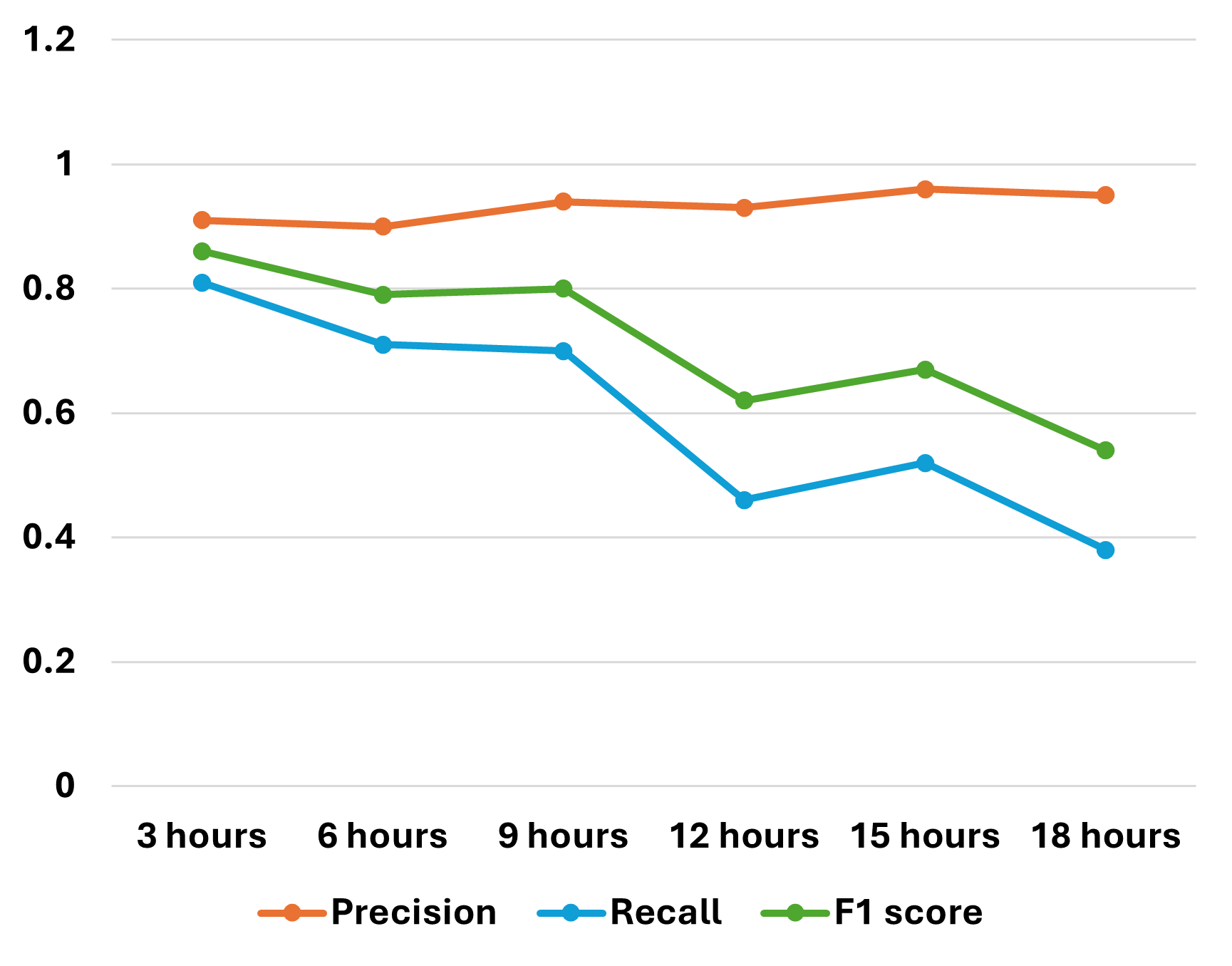}
    \caption{Reducing recall over time}
    \label{fig:Recall}
\end{figure}

In practical fall detection scenarios, as users continuously interact with the system, the model predominantly receives feedback data, which is mostly ADL. During personalized retraining, we combine the original training dataset with the user's newly collected feedback data. Over time, this naive approach progressively increases the data imbalance toward ADLs, inadvertently introducing bias into the model. We used a subset of the SmartFallMM~\cite{smartfallMM} dataset as the initial training data, which includes fall and ADL samples from 30 participants (a total of 1,950 trials). To personalize the model for each user, we merged their collected feedback data with this dataset and trained a new model tailored to that individual. Figure~\ref{fig:Recall} illustrates this effect for a particular $i$-th user. Initially, retraining the model with just three hours of the user's ADL data yielded a recall of 0.81. However, with each subsequent retraining session (each adding an additional three hours of ADL data), the recall steadily decreased. By the sixth retraining cycle, recall dropped below 0.50, demonstrating a notable reduction in the model's capacity to accurately detect falls. On the other hand, precision improved incrementally with additional ADL data. This indicates that while the model generates fewer false positives over time, it simultaneously loses effectiveness in identifying actual fall events.


Our systematic analysis shows that incorporating more than 20\% feedback data during personalization results in a noticeable recall drop on the SmartFallMM dataset, indicating increased bias toward ADL patterns. This highlights the risk of overfitting to ADLs when fall data are scarce. To address this issue, we investigate strategies that maintain recall across successive personalization rounds.

%% file: sections/related_work.tex
\section{Related Work}

Personalization remains a key challenge in human activity recognition due to large inter‑user variability and limited labeled data. Recent work addresses this through \textit{adaptive modeling}, which customizes activity recognition for each individual by learning user-specific motion patterns and incorporating user feedback, \textit{unsupervised or self‑supervised learning}, which learns representations from unlabeled data, and \textit{active learning}, which strategically selects informative samples for labeling.

\paragraph{Adaptive Modeling}
Lingmei and Weisong~\cite{Chameleon2016} introduced a subject‑adaptive thresholding mechanism using a waist‑worn sensor and a weighted combination of user‑specific and group‑level thresholds. While this approach better captures individual variability, it relies heavily on heuristic rules and hand‑crafted features. Vallabh et al.~\cite{pers2020vallabh} proposed a personalized fall detector using smartphone accelerometer data and angle‑based outlier detection without explicit fall data, but its unbounded retraining risks overfitting and drift, and its signal magnitude vector‑based  (SMV)segmentation is sensitive to noisy, low‑fidelity sensors. Brian et al.~\cite{US12027035B2} similarly used subject‑specific non‑fall distributions and a fixed threshold from external fall data, refined by user feedback, but their method assumes stable sensor quality and lacks online adaptation, limiting robustness in dynamic environments.

Ferrari et al.~\cite{ferrari2020personalization} improved accuracy by exploiting user similarity based on physical traits and sensor signals, yet their approach is less suitable for fall detection, where metadata are often unavailable, and fall events are rare and highly individual. Ngu et al.~\cite{NguIJNS2022} proposed a real‑time personalized system using an LSTM‑based model deployed on wrist‑worn smartwatches, leveraging user feedback for periodic retraining from scratch. They trim feedback segments around acceleration spikes to reduce noise, but this trimming is purely temporal and does not account for semantic informativeness, so it cannot distinguish high‑quality from redundant samples. 

Diraco et al.~\cite{diraco2023human} highlighted the importance of personalization and context in smart living environments, but their work focuses on multi‑sensor‑based systems rather than single‑device wearables. Mhalla et al.~\cite{mhalla2024domain} introduced a domain adaptation framework that personalizes models without labeled data. However, the method assumes a smooth domain shift and continuous data streams, conditions that rarely hold in sporadic, event‑driven fall detection. Our approach is tailored for fall detection on resource‑constrained wearables, operating without additional user metadata or external sensors.

\paragraph{Unsupervised and Self‑supervised Learning}
Unsupervised and self‑supervised techniques have been increasingly used to reduce dependence on labeled data. 
For example, Li et al.~\cite {li2025unsupervised} proposed a self-supervised framework that enhances skeleton-based action recognition through feature enrichment and fidelity preservation. Qin et al.~\cite{qin2025progressive} introduced a progressive semantic learning approach that incrementally refines ambiguous action representations in an unsupervised manner. On the other hand, Xu et al.~\cite{xu2025positive} addressed a core limitation in contrastive learning, excessive intra-class variance, by introducing a fuzzy threshold mechanism to better distinguish between positive and negative samples based on similarity distributions using multi-perspective skeleton data.
These methods excel on vision‑based skeleton data but are less suited to wearable sensor‑based fall detection, where irregular sampling, high noise, and extreme class imbalance dominate. They also lack mechanisms for temporal distortions, rare‑event modeling, and online personalization.

\paragraph{Active Learning}
Active learning methods aim to maximize model performance while minimizing labeling costs by strategically selecting the most informative unlabeled samples for annotation. For example, Hasan and Roy‑Chowdhury~\cite{7410873} employed entropy-based active learning with conditional random fields to select informative samples for video-based activity recognition. Bi et al.~\cite{9153742} combined uncertainty, diversity, and representativeness criteria to select informative samples and identify new activity classes. Kazllarof and Kotsiantis~\cite{Kazllarof2022} applied active learning with least confidence sampling to time series feature extraction across spectral, statistical, and temporal domains. More recently, Arrotta et al.~\cite{arrotta2023} introduced SelfAct, which combines self‑supervised learning with an unsupervised active learning strategy based on cluster density, selecting representative samples from clusters of embeddings generated by a pre‑trained model. 

Our framework addresses a related but distinct challenge. Active learning assumes control over which samples to present for labeling, which is impractical in real-world fall detection, where alerts occur naturally during daily activities, and users can only confirm or reject them as they happen. Rather than selecting which samples to label, we operate on already-labeled user feedback (TP/FP confirmations) and focus on selecting which confirmed samples to incorporate during retraining. This is necessary because feedback data is inherently imbalanced, i.e., false positives from routine ADLs vastly outnumber true fall events, biasing the model toward non-fall patterns and reducing fall sensitivity. Our clustering-based gradient selection addresses this by identifying diverse, informative examples from the feedback pool to balance class representation.

%% file: sections/method.tex
\section{Methodology}
\label{sec:method}

\begin{figure}[t!]   
    \centering
    \includegraphics[width=.65\textwidth]{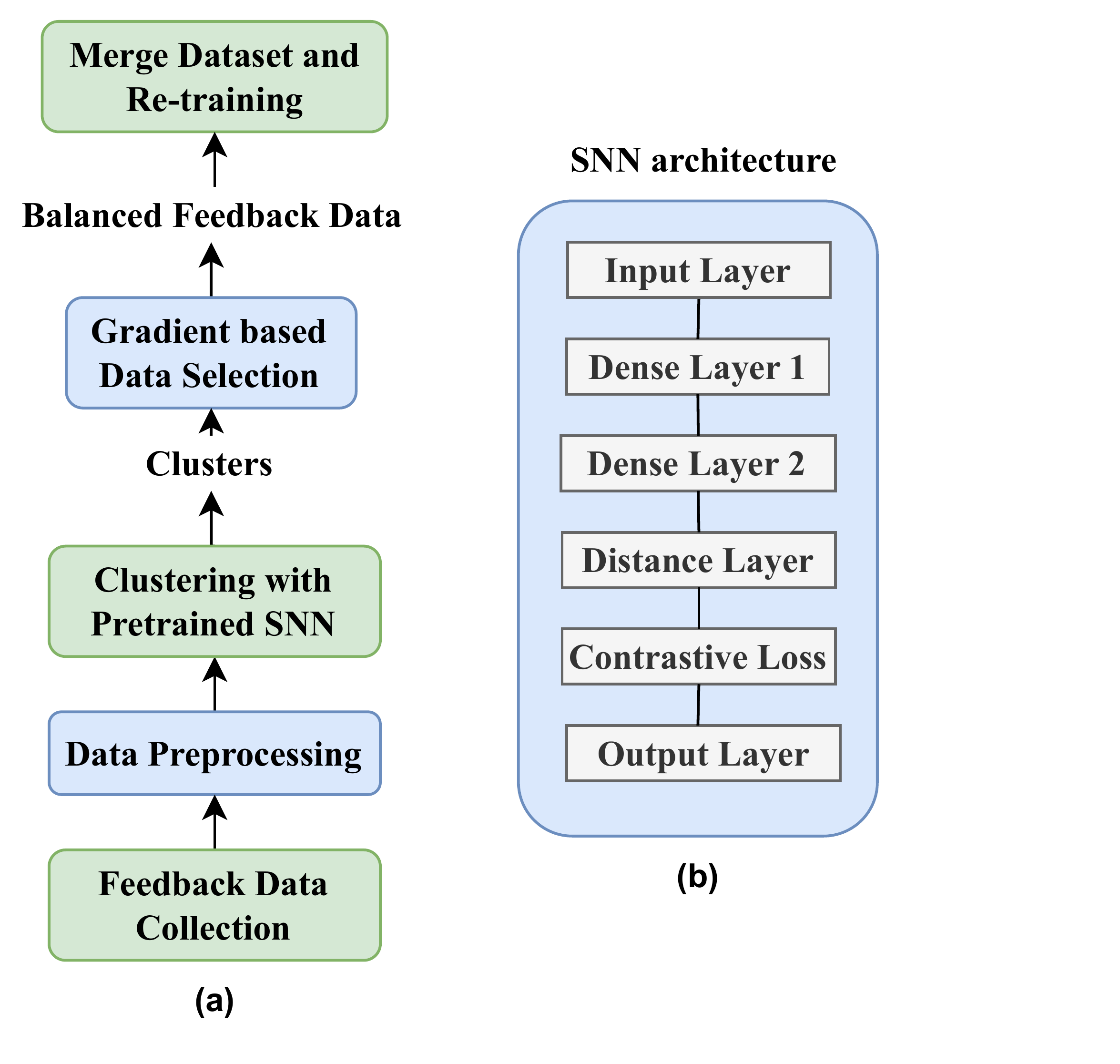}
     \caption{(a) Overview of the pipeline for training personalized models ($M_{Pi}$) \& (b) SNN architecture}
     \label{fig: SNN}
 \end{figure}

Figure~\ref{fig: SNN} presents an overview of the personalization pipeline for each user. The pipeline begins with the feedback data collected from the user while using a fall detection model. For this, we used a pretrained transformer model ($M_O$) trained with a subset of the SmartFallMM dataset \cite{smartfallMM}, deployed on a smartwatch for fall detection. Whenever the model triggers fall alerts as users go about their daily activities, the system 
prompts the user to provide feedback by confirming whether the alert was a true positive (TP), an actual fall, or a false positive (FP), an ADL incorrectly classified as a fall. 

All confirmed FP and TP are automatically archived to the cloud. False positives are especially useful for helping the model learn to distinguish between fall-like ADLs and actual falls. True positives are equally important, as they represent real fall patterns specific to each user. These labeled samples (both FP and TP) are stacked and clustered using a pretrained Siamese Neural Network (SNN), 
a general-purpose model trained in advance using the same subset of the SmartFallMM dataset as the initial transformer model $M_O$. 
The SNN is stored in the cloud and reused across all users
to consistently group similar feedback samples.
Finally, to build a compact and representative training set, we compute the gradient for each sample and select the top $x$ samples from each cluster based on their contribution to model learning. These selected instances, representing diverse and challenging examples, are then combined with the original training data used by $M_O$ to retrain a personalized fall detection model $M_{Pi}$. 

\subsection{Feedback Data Collection}

We deployed a fall detection model ($M_O$) 
on a smartwatch-based fall detection application (details omitted for anonymity) and recruited ten participants to use the App at home to collect feedback data. 

\subsubsection{Fall Detection Model}
\label{sec:transformer}
\begin{figure}[t]
\centering
    \includegraphics[width=0.65\linewidth]{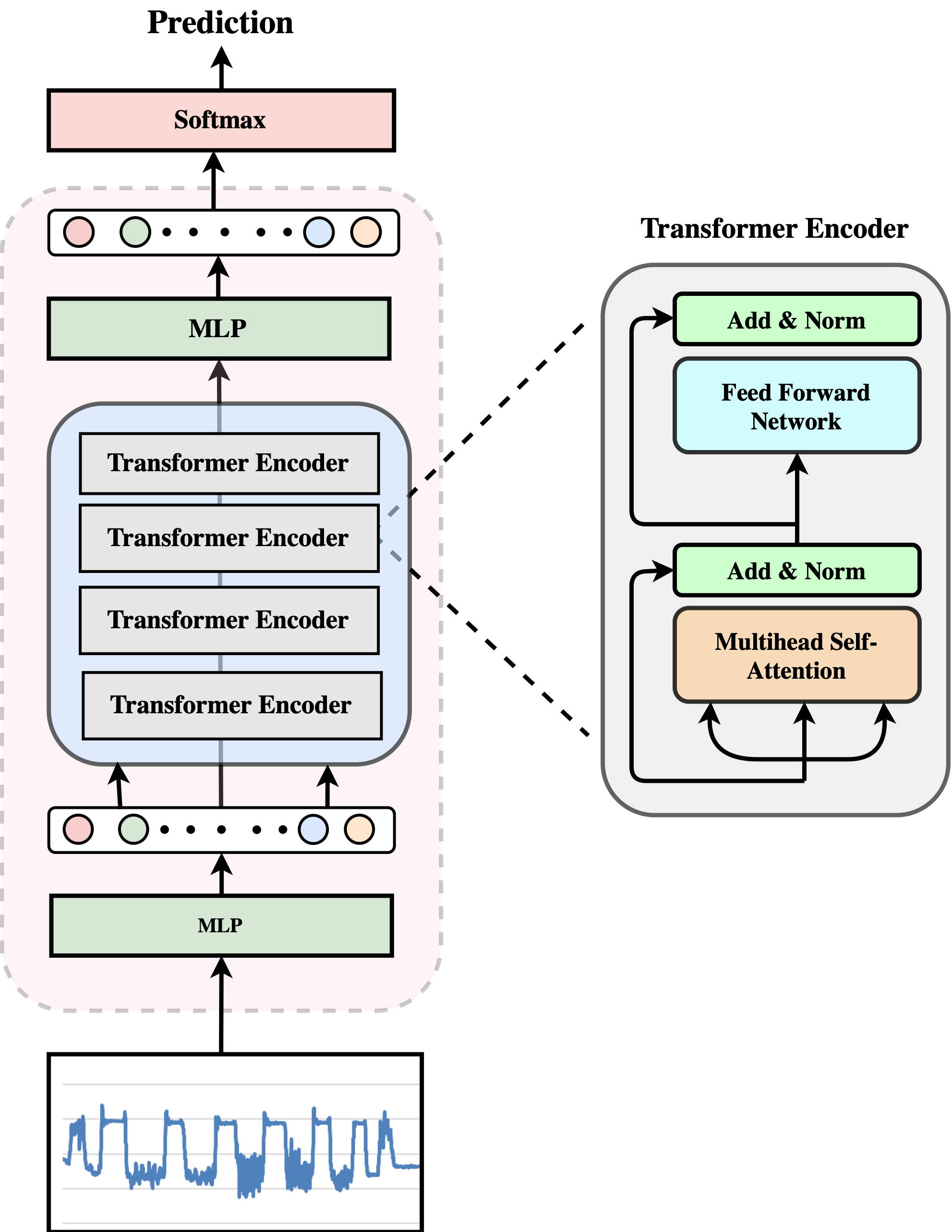}
    \caption{Transformer architecture}
    \label{fig:Transformer}
\end{figure}

The initial fall detection model, $M_O$ was obtained by training a Transformer model~\cite{s24196235} with a subset of the SmartFallMM dataset. 
Our implementation used a Transformer architecture, illustrated in figure \ref{fig:Transformer}, comprising 4 encoder layers and 4 multi-head attention heads, with an attention embedding size of 128 and a dropout rate of 0.25. The final representation was passed through a multi-layer perceptron (MLP) with 1, 8, and 16 layer sizes. We applied a sigmoid activation function at the output and trained the model using binary cross-entropy loss. Training was conducted over 100 epochs using the Adam optimizer, with a batch size of 64 and a learning rate of 0.001. To promote subject-independent generalization, we adopted a leave-one-subject-out cross-validation strategy for evaluation.

\subsubsection{Data Collection via User Interaction}
Once the smartwatch is paired with a smartphone, data sensing is activated through the phone's App interface. After activation, all interactions with the users occur through the smartwatch. The App continuously senses 128 accelerometer data points from the wrist's motion before making a fall prediction. Each data point was sampled at an interval of 32 ms. If a fall is detected, the watch will prompt the user for confirmation. A “NO” response is recorded as a false positive (FP), while a “YES” response is followed by an additional prompt asking if help is needed. Confirmed falls, whether help is requested or not, are saved as true positives (TP). This labeled feedback data is periodically uploaded to the cloud server and will be used to train the personalized model $M_{Pi}$ for each user.

In our experimental studies, each user will use the system 
for three hours per day over six days. During this time, the user’s only task is to provide feedback by confirming whether any fall alert triggered by the model is a true positive or a false positive.
At the end of the sixth day, the saved feedback data is downloaded from the cloud and clustered. This enables the identification of different clusters of ADL activities and selection of informative samples to personalize the model for each user.

\subsection{Data Preprocessing}

We start the personalization process by segmenting each feedback sample using a sliding window approach with overlap. Through experimentation, a window size of 128 data points with a 10-point overlap provided optimal results for distinguishing falls from ADLs. With the smartwatch sampling rate of 32~ms, each window represents approximately 4~s of activity. For each window, we compute similarity scores using three complementary measures: Pearson’s Linear Correlation Coefficient (PLCC)~\cite{benesty2009pearson}, Dynamic Time Warping (DTW)~\cite{Turan2016-sensor}, and Cosine Similarity~\cite{ngoc2023cosine}. These metrics jointly capture aligned temporal trends, temporal misalignments, and complex feature-space relationships, enabling robust and discriminative clustering of user feedback data.

PLCC quantifies the linear relationship between two sequences, identifying patterns that move together over time, even if their scales differ. It is defined as:
\[
r_{x,y} = \frac{\sum_{i=1}^{n} (x_i - \overline{x})(y_i - \overline{y})}{\sqrt{\sum_{i=1}^{n} (x_i - \overline{x})^2} \times \sqrt{\sum_{i=1}^{n} (y_i - \overline{y})^2}}
\]
where \(x_i\) and \(y_i\) are data points, and \(\overline{x}\) and \(\overline{y}\) are their means.

DTW measures the optimal alignment between two sequences that may differ in speed or duration—useful in fall scenarios where timing varies. It is computed as:
\[
DTW(i, j) = d(x_i, y_j) + \min \begin{cases}
DTW(i-1, j), \\
DTW(i, j-1), \\
DTW(i-1, j-1)
\end{cases}
\]
where \(d(x_i, y_j)\) is the distance between points.

Cosine Similarity evaluates the angle between two vectors, ignoring magnitude, making it suitable for deep feature comparisons:
\[
\cos(\theta) = \frac{\mathbf{x} \cdot \mathbf{y}}{\|\mathbf{x}\| \times \|\mathbf{y}\|}
\]

\subsection{Clustering with Pretrained SNN}
We first trained a Siamese Neural Network (SNN) \cite{shalaby2021data} on the SmartFallMM dataset \cite{smartfallMM} and used the pretrained model to cluster feedback data based on similarity to labeled ADL patterns. SNNs are well-suited for this task because they learn pairwise similarity and perform effectively with limited labeled data, making them particularly useful when few samples are available per class \cite{shalaby2021data}. By producing discriminative embeddings, the SNN enables semi-supervised clustering and identification of high-confidence unlabeled ADL samples.

The SNN architecture used for clustering consists of three main components: a base network, a distance layer, and a contrastive loss function, as illustrated in Figure~\ref{fig: SNN}(b).  

\textbf{1. Base Network:}  
The base network serves as a shared feature extractor that transforms each input vector into a fixed-length embedding. Both inputs pass through identical networks with shared weights to ensure consistent feature extraction.  
The architecture includes four layers: 

\begin{itemize}
    \item \textbf{Input Layer:} Feature vector of shape (9,), representing similarity metrics (cosine similarity, PLCC, DTW) between accelerometer axes (x, y, z).  
    \item \textbf{Dense Layer 1:} 128 neurons with ReLU activation.  
    \item \textbf{Dense Layer 2:} 64 neurons with ReLU activation.  
    \item \textbf{Output Layer:} 32 neurons with linear activation, producing the embedding vector.  
\end{itemize}

\textbf{2. Distance Layer:}  
This layer computes the similarity between two embeddings from the base network using the Euclidean (L2) distance:  
\[
\text{Distance} = \left\| \text{Embedding}_A - \text{Embedding}_B \right\|_2
\]
A smaller distance indicates higher similarity, while a larger distance indicates lower similarity.

\textbf{3. Contrastive Loss:}  
Training is guided by the contrastive loss function~\cite{NEURIPS2021_27debb43}, which minimizes distances for similar pairs (\(y = 1\)) and maximizes them for dissimilar pairs (\(y = 0\)) up to a margin:
\[
L = y \cdot (\text{distance})^2 + (1 - y) \cdot \max(0, \text{margin} - \text{distance})^2
\]
where \( y \in \{0, 1\} \) and the margin is a tunable parameter (e.g., 1.0).

The SNN was trained using data from 30 participants in the SmartFallMM dataset~\cite{smartfallMM} for 100 epochs. This approach allows the model to embed input signals in a discriminative feature space, effectively grouping similar motion patterns into clusters for use in the selective feedback process.

\subsubsection{Clustering}


Once the pretrained SNN model is obtained, each feedback instance $x_i \in X$ is passed through it to generate its 
embedding, forming a low-dimensional feature space in which similar motion patterns are grouped for clustering and subsequent selective data selection.
After that, the DBSCAN clustering algorithm \cite{khan2014dbscan} is employed on the embeddings to group them into clusters as: 
\begin{equation}
    \mathcal{C} = \{C_1, C_2, \dots, C_K\} = \text{DBSCAN}\!\left(f_{\text{SNN}}(X)\right)
\end{equation}
where $f_{\text{SNN}}(\cdot)$ denotes the SNN model, 
and $\mathcal{C}$ represents the resulting set of clusters. Since the SNN has learned to place similar items close together, the clusters correspond to the true classes. 
Unlike $K$-means clustering, DBSCAN automatically determines the appropriate number of clusters based on data density, ensuring that clusters are formed only when sufficient similar samples exist, and thereby preventing the creation of artificial or redundant groups.




\subsection{Gradient-based Data Selection}
After obtaining the $\mathcal{C}$ clusters, we calculate the gradient for each window of data within each cluster $C_j$ to identify the windows exhibiting the most significant temporal changes. Since the data points within a window are uniformly spaced in time, we compute the average gradient by taking the mean of the consecutive differences between points. The gradient for a window $w_i \in C_j$ with $n$ points is computed as~\cite{JMLR:v17:13-351}:
\begin{equation}
   g_i = \frac{1}{n-1} \sum_{z=1}^{n-1} \left| y_{z+1}^{(i)} - y_{z}^{(i)} \right| 
\end{equation}
where $y_z^{(i)}$ represents the signal value at the $z$-th time step of the $i$-th window. A higher $g_i$ value indicates greater temporal variation within the window, reflecting more dynamic motion patterns such as falls or abrupt limb movements.

Once the gradient values are computed, we select the windows with the highest $g_i$ values, as they contribute most to model learning. Empirical evaluation on the SmartFallMM dataset showed that using about one-fifth of the data achieves an optimal trade-off between accuracy and efficiency. Accordingly, $x = 0.2 \times \frac{|X|}{|\mathcal{C}|}$ samples are allocated per cluster to provide sufficient representation of user-specific motion patterns without redundancy. Clusters with fewer than $x$ samples include all available data, with the remaining quota filled from other clusters, while those with more than $x$ samples apply gradient-based selection to retain only the most informative windows, yielding a balanced and diverse personalized dataset.

\subsection{Merge Dataset For Retraining and Evaluation}
The selected feedback samples $\mathcal{D}_{\text{sel}} = \bigcup_{j=1}^{|\mathcal{C}|} S_j$, where $S_j = \operatorname{Top}_{x}\!\left(\{g_i \mid x_i \in C_j\}\right)$, are then merged with the original SmartFallMM dataset $\mathcal{D}_{\text{orig}}$ to create a retraining dataset $\mathcal{D}_{\text{merged}} = \mathcal{D}_{\text{orig}} \cup \mathcal{D}_{\text{sel}}$. 
To assess the effectiveness of the proposed selective personalization approach, we evaluated it under three standard retraining strategies: Training from Scratch (TFS), Transfer Learning (TL), and Few-Shot Learning (FSL), yielding personalized models $M_{Pi}$. 
These strategies are employed to examine whether the selectively merged dataset $\mathcal{D}_{\text{merged}}$ improves model performance across different retraining paradigms.

To train and evaluate personalized models ($M_{P_i}$), each participant’s feedback data were split using an 80--20 within-subject partition. For each participant, the base model ($M_0$) was retrained using 80\% of the feedback data combined with the original training subset ($\mathcal{D}_{\text{orig}}$) of SmartFallMM, producing a participant-specific model. The remaining 20\% of feedback data were merged with the test subset, containing both fall and ADL events, for evaluation. This split balances effective retraining and reliable evaluation without overfitting. As a result, each participant yields three personalized models corresponding to TFS, TL, and FSL, producing a total of 30 personalized models for ten participants.

%% file: sections/experimental_design.tex
\section{Experimental Design}

We used the Transformer model described in Section \ref{sec:transformer} as our initial model ($M_O$). This model was first trained on a subset of the SmartFallMM dataset \cite{smartfallMM} and then deployed to collect user feedback data during real-world use. 
The collected feedback from each participant was later used to retrain the model, following three strategies: Training From Scratch (TFS), Transfer Learning (TL), and Few-Shot Learning (FSL).

\subsection{Initial Model Training and Feedback Data Collection}


The initial Transformer model ($M_0$) was trained on the SmartFallMM dataset \cite{smartfallMM}, which contains motion data from 30 participants performing 13 activities, including five simulated falls and eight ADLs. The trained model($M_0$) was deployed on WearOS smartwatches paired with Android phones. Ten participants used the system for six days, during which accelerometer data were continuously collected and segmented into 128-sample windows for real-time fall detection. Users provided feedback on detected falls, which was stored in Couchbase to create personalized datasets of confirmed fall and non-fall events. These datasets were subsequently used to retrain personalized models under identical experimental conditions.

All models were implemented in TensorFlow and trained on a Dell Precision 7820 workstation with an NVIDIA GeForce GTX 1080 GPU. Each experiment was repeated three times, and results were averaged for consistency. Hyperparameters such as learning rate, batch size, embedding dimension, and attention head count were empirically tuned for each approach, and all models were trained and evaluated using identical dataset splits and computational settings to ensure fair comparison.

\subsection{Retraining Approaches}
After six days of use, the collected feedback data were retrieved and merged with the original SmartFallMM subset to create personalized datasets for each participant. These datasets were used to retrain models using TFS, TL, and FSL, with all experiments conducted under identical hardware and computational conditions for fair comparison.

\subsubsection{Training from Scratch}

In this approach, the personalized models $M_{Pi}$ for each user $i$ are obtained by retraining the $M_O$ model using each participant’s combined dataset $\mathcal{D}_{\text{merged}}$, which included both the SmartFallMM subset and their individual selective feedback data.
This process allowed the model to completely relearn from both general and user-specific data distributions without relying on pretrained parameters, producing $M_{TFS}$ personalized model.  
This approach required the longest training time and the greatest computational cost, as all model parameters were optimized from initialization. While highly effective in terms of performance, the high runtime makes this method less practical for on-device retraining or frequent updates.

\subsubsection{Transfer Learning}
For the Transfer Learning (TL) approach, we applied the fine-tuning method described by Maray et al. \cite{kuhar}, adapting it to the Transformer-based architecture illustrated in Figure~\ref{fig:Transformer}. The lower encoder layers, responsible for extracting fundamental motion features, were frozen to retain general motion dynamics learned from the initial training. The upper layers and the classification head were then retrained using each participant’s selective feedback data. The model parameters were optimized using the Adam optimizer with categorical cross-entropy loss, and hyperparameters such as learning rate, embedding dimension, and number of attention heads were empirically tuned on $\mathcal{D}_{\text{merged}}$ dataset, producing $M_{TL}$ personalized model.

\subsubsection{Few Shot Learning}
The Few Shot Learning (FSL) approach was implemented following the method of Zhou et al. \cite{10.1145/3442381.3450006}, designed to enable rapid personalization with limited feedback data. The model shared the same Transformer-based backbone as the TL model, consisting of an input embedding layer, positional encodings, stacked Transformer encoder blocks, and a fully connected classification head. The lower layers were kept frozen to preserve general temporal representations, while only the upper layers and the classifier were fine-tuned using a small set of feedback samples.
Unlike the TFS and TL strategies, which retrain or fine-tune the model using the entire merged dataset $\mathcal{D}_{\text{merged}}$, the FSL model uses only a small subset, resulting in a few-shot-based personalized model $M_{FS}$.

%% file: sections/results.tex
\section{Results and Analysis}

\label{sec:results}
\subsection{Model Performance}
\subsubsection{Offline Performance}

\begin{figure}[t!]
\centering
    \includegraphics[width=.7\textwidth]{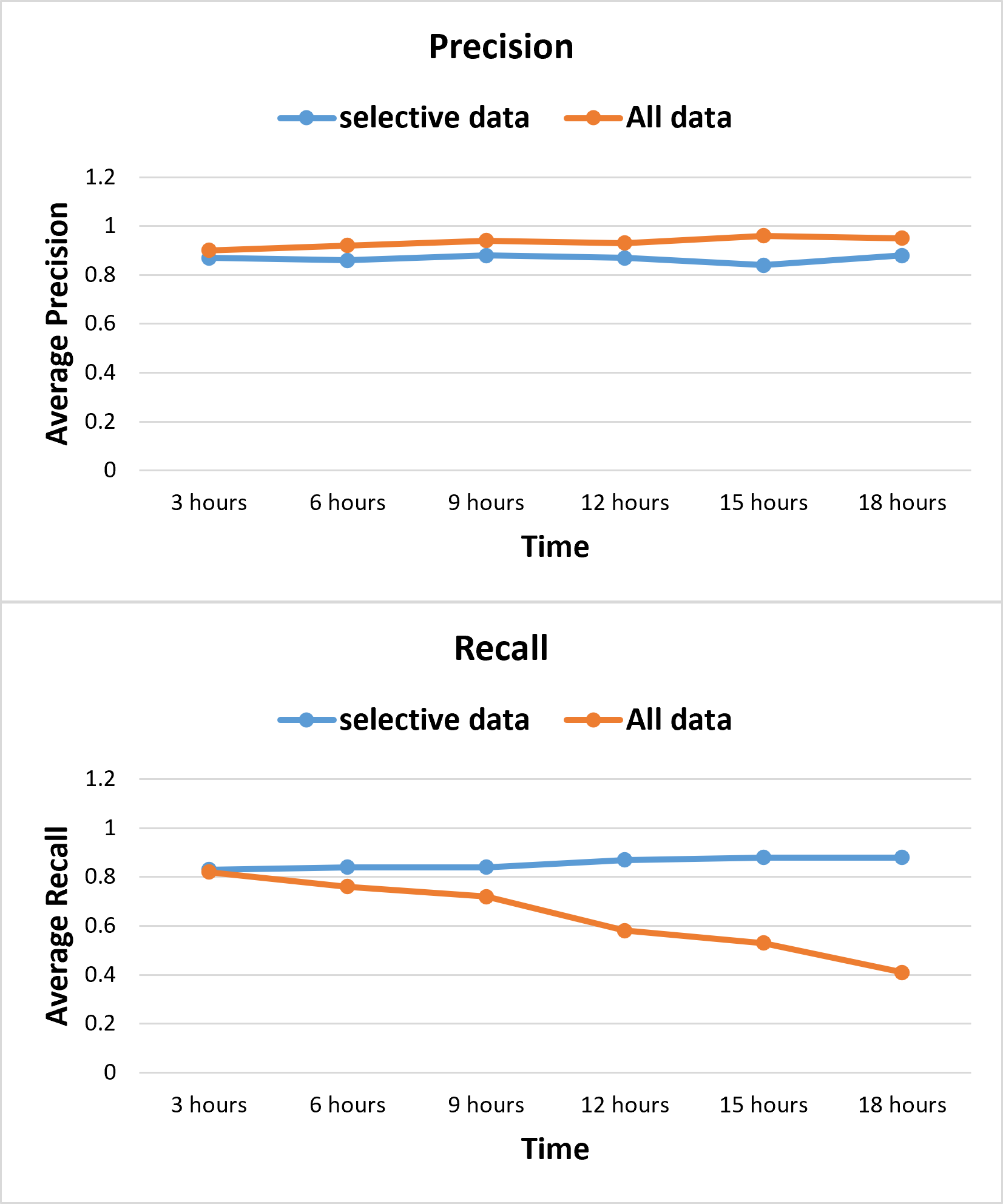}
    \caption{Precision and recall over time for models trained with selective data vs. all feedback data}
    \label{fig:Result_compare} 
\end{figure}

Figure~\ref{fig:Result_compare} shows the average result from 10 participants using the test set that was built from 20\% of that user's feedback data with the TFS retraining strategy. We can see how precision and recall evolve for different personalization strategies: using selectively chosen data versus all feedback data.

For precision (top panel), both strategies maintain consistently high values over time. The all-data approach yields slightly higher precision but at the cost of bias, i.e., by including many repetitive ADL samples, the model becomes overly conservative, reducing false positives but missing true falls. In contrast, the selective-data method, though marginally lower in precision, achieves a better balance by preserving sensitivity to both ADL and fall events.

For recall (bottom panel), the difference is more pronounced. Recall with selective data increases steadily from about 0.80 to 0.88, while in the all-data condition, it declines sharply from 0.80 to 0.41 as feedback accumulates. The unfiltered data led to overfitting on ADL and poor fall sensitivity. Averaged across participants, the selective-data strategy reached an F1-score of 0.88 (compared to 0.82 for $M_{TFS}$ and 0.48 for the all-data case), confirming that curated, high-gradient samples yield more generalizable and reliable models for personalized fall detection.

\subsubsection{Real Time Performance}


\begin{table}[t!]
\centering
\caption{Real-time evaluation results of 10 participants using all 3 approaches of retraining vs {$M_O$} (bold values indicate the highest F1 score achieved across models for a given participant)}
\label{tab: realtime}
\scalebox{1}{
\begin{tabular}{|c|c|ccc|}
\hline
\multirow{2}{*}{\textbf{Participant ($i$)}} & \textbf{Baseline} & \multicolumn{3}{c|}{\textbf{Personalized Models ($M_{Pi}$)}}                                         \\ \cline{2-5} 
                                            & \textbf{$M_O$}    & \multicolumn{1}{c|}{\textbf{$M_{TFS}$}} & \multicolumn{1}{c|}{\textbf{$M_{TL}$}} & \textbf{$M_{FS}$} \\ \hline
Participant 1                               & 0.69              & \multicolumn{1}{c|}{0.89}               & \multicolumn{1}{c|}{0.72}              & 0.79              \\ \hline
Participant 2                               & 0.71              & \multicolumn{1}{c|}{0.9}                & \multicolumn{1}{c|}{0.69}              & 0.78              \\ \hline
Participant 3                               & \textbf{0.76}     & \multicolumn{1}{c|}{0.92}               & \multicolumn{1}{c|}{0.67}              & 0.8               \\ \hline
Participant 4                               & 0.68              & \multicolumn{1}{c|}{0.93}               & \multicolumn{1}{c|}{0.7}               & 0.76              \\ \hline
Participant 5                               & 0.73              & \multicolumn{1}{c|}{\textbf{0.94}}      & \multicolumn{1}{c|}{0.75}              & 0.74              \\ \hline
Participant 6                               & \textbf{0.76}     & \multicolumn{1}{c|}{0.88}               & \multicolumn{1}{c|}{0.68}              & 0.77              \\ \hline
Participant 7                               & 0.75              & \multicolumn{1}{c|}{0.88}               & \multicolumn{1}{c|}{0.65}              & 0.81              \\ \hline
Participant 8                               & 0.75              & \multicolumn{1}{c|}{\textbf{0.94}}      & \multicolumn{1}{c|}{0.71}              & 0.76              \\ \hline
Participant 9                               & 0.71              & \multicolumn{1}{c|}{0.93}               & \multicolumn{1}{c|}{\textbf{0.77}}     & \textbf{0.82}     \\ \hline
Participant 10                              & 0.74              & \multicolumn{1}{c|}{0.92}               & \multicolumn{1}{c|}{0.73}              & 0.79              \\ \hline
Average                                     & 0.73              & \multicolumn{1}{c|}{\textbf{0.91}}      & \multicolumn{1}{c|}{0.71}              & 0.78              \\ \hline
\end{tabular}}
\end{table}

Table~\ref{tab: realtime} summarizes the F1-scores achieved by each participant for the baseline Transformer ($M_{O}$) and three retraining strategies: Training from Scratch ($M_{TFS}$), Transfer Learning ($M_{TL}$), and Few-Shot Learning ($M_{FS}$). Selective retraining consistently improves performance across all participants. The baseline model ($M_{O}$) achieved an average F1-score of 0.73, while retraining from scratch ($M_{TFS}$) yielded the best performance with an average F1-score of 0.91, representing an improvement of approximately 25\%.

Few-shot learning ($M_{FS}$) also performed well, achieving an average F1-score of 0.78 and offering a favorable balance between accuracy and efficiency. In contrast, transfer learning ($M_{TL}$) underperformed with an average F1-score of 0.71, indicating limited adaptability to user-specific motion patterns. For example, Participant~4’s F1-score improved from 0.68 ($M_{O}$) to 0.93 ($M_{TFS}$), while Participant~7’s increased from 0.75 to 0.88. Overall, these results confirm that selective data personalization enhances model robustness, with TFS achieving the highest accuracy and FSL providing a practical trade-off between performance and computational efficiency.

\subsection{Performance of SNN}

\begin{table}[t!]
\centering
\caption{Accuracy of SNN on Different Datasets}
\label{tab:Dataset-result}
    \begin{tabular}{|c|c|}
    \hline
    \textbf{Dataset} & \textbf{Accuracy} \\ \hline
    SmartFallMM      & 91\%                     \\ \hline
    HAR              & 94\%                     \\ \hline
    CZU-MHAD         & 85\%                   \\ \hline
    K-Fall           & 86\%                  \\ \hline
    \end{tabular}
\end{table}

To ensure the accuracy of SNN for our clustering of ADL data, we compare the performance of the proposed SNN across four different datasets.
Table~\ref{tab:Dataset-result} summarizes the accuracy of the SNN across four benchmark datasets: SmartFallMM, HAR, CZU-MHAD, and K-Fall. Accuracy was determined by assigning each cluster to the class with the most representative samples. The model demonstrates strong generalization ability across a range of human activity recognition and fall detection tasks, with accuracy values ranging from 85\% to 94\%.

The highest performance is observed on the HAR \cite{morales2023human} dataset, where the model achieves an accuracy of 94\%. This dataset contains clean, structured smartphone sensor data of routine human activities performed in controlled settings. 

On the SmartFallMM dataset, which is the primary dataset used for model development and evaluation, the model achieves an accuracy of 91\%. This dataset is characterized by real-world, unconstrained activity data collected in diverse conditions. 

The model also performs well on the CZU-MHAD \cite{chao2022czu} dataset with an accuracy of 85\%. This dataset includes multimodal motion data captured from multiple users performing a broader set of actions. 

The K-Fall dataset \cite{yu2021large} achieved a strong clustering accuracy of 86\%, which is especially notable because it does not include wrist-based sensor data. Unlike other datasets, which are collected from a smartwatch on the wrist, K-Fall uses a sensor placed on the lower back. This means the motion patterns are quite different.

The SNN model achieves consistently high performance across all datasets, demonstrating its adaptability to different sensors and activities. These results confirm the proposed method’s potential to be used as a semi-supervised clustering method for feedback data.

\begin{figure*}[t!]
\centering
     \includegraphics[width=1\columnwidth]
    {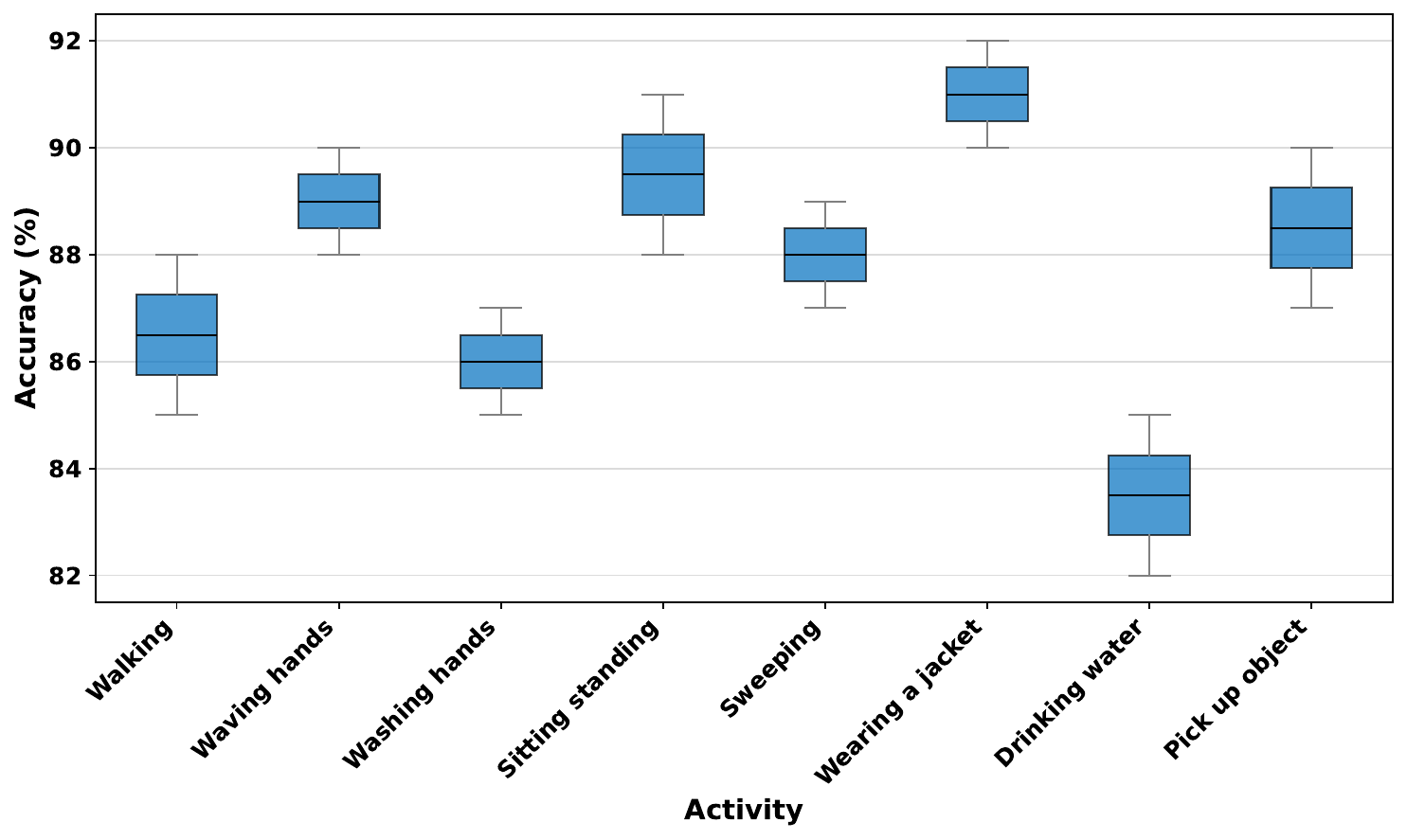}
    \caption{Clustering accuracy for each activity}
    \label{fig:accuracy}
\end{figure*}

In addition to model-level evaluation, we assessed the accuracy of the clustering mechanism used for selective data sampling only on SmartFallMM. This component identifies high-gradient, informative segments from user feedback, enabling efficient and robust retraining by emphasizing representative motion patterns.

Figure~\ref{fig:accuracy} shows that clustering accuracy varies slightly across ADLs, with an overall average of 91\%. Activities with distinct and consistent motion patterns, such as Wearing a jacket and Sitting and standing, achieved the highest accuracies (92\% and 91\%, respectively), while Waving hands and Pick up object both reached 90\%, indicating effective separation of upper-limb activities.

Conversely, activities involving smaller or more repetitive wrist movements, such as Washing hands (87\%) and Drinking water (85\%), exhibited comparatively lower clustering accuracies. This reduction can be attributed to overlapping motion characteristics and limited dynamic variation, which make these activities more challenging to distinguish in the feature space. Despite these minor variations, the clustering accuracy across all activities remained consistently above 85\%, indicating the reliability and stability of the clustering method.

Overall, the results depicted in Figure \ref{fig:accuracy} confirm that the SNN-based clustering framework effectively identifies representative and informative motion segments for selective retraining. By achieving an average clustering accuracy of 89\%, the process ensures that feedback-driven fine-tuning concentrates on the most meaningful data samples.

\subsection{Ablation Study}
\subsubsection{Similarity Metrics vs. Basic Statistical Features}
To assess the choice of similarity metrics, we conducted an ablation study. Table \ref{tab:ablation} presents the performance of three different personalization strategies regarding precision, recall, and F1-score. For this evaluation, we ran the full personalization pipeline using a different instance of the SNN for each experiment to ensure robustness and consistency in the results. The method utilizing all 3 similarity metrics, such as cosine similarity, PLCC, and DTW, achieves the highest overall performance, with a precision of 0.92, a recall of 0.9, and an F1-score of 0.91. 

In contrast, substituting these similarity metrics with basic statistical features (minimum, maximum, mean, and standard deviation) results in significantly lower performance. The F1-score drops to 0.63, with a significant decline in recall (0.55), suggesting that simple descriptive statistics did not sufficiently capture the complex patterns necessary for accurate fall detection.

Similarly, using the raw time-series data without engineered similarity or statistical features leads to a slightly better recall (0.72) than the statistical feature method. Still, the overall F1-score remains low at 0.67 due to a notable drop in precision (0.63). This underscores the importance of carefully selecting features that reflect meaningful relationships in time-series data for effective personalization.


\begin{table}[t!]
\centering
\caption{Comparing the performance of SNN with different inputs}
\label{tab:ablation}
    \begin{tabular}{|c|c|c|c|}
    \hline
    \textbf{Input for SNN}                                                                      & \textbf{Precision} & \textbf{Recall} & \textbf{F1-Score} \\ \hline
    Similarity metrics                                                                          & 0.92               & 0.9            & 0.91              \\ \hline
    \begin{tabular}[c]{@{}c@{}}Basic statistical features \\ (min, max, mean, std)\end{tabular} & 0.75               & 0.55            & 0.63              \\ \hline
    Raw time-series data                                                                       & 0.63               & 0.72            & 0.67              \\ \hline
    \end{tabular}
\end{table}

\subsubsection{Gradient based vs Random Data Selection}
To evaluate the effect of gradient-based (GB) selection, we compared the full pipeline using GB sampling with random selection. After clustering, random selection of 20\% samples per cluster achieved an average F1-score of 0.85 (0.80-0.91 across runs), whereas GB selection, which prioritizes high-gradient windows, improved the average F1-score to 0.91, demonstrating its effectiveness in identifying informative feedback samples.

We further examined the impact of feedback size on personalization. As shown in Table~\ref{tab:fd}, performance increased from an F1-score of 0.80 at 5\% feedback to 0.91 at 20\%, before slightly declining to 0.89 at 25\% due to increased redundancy and reduced data diversity. Overall, selecting approximately 20\% of feedback data using GB sampling provides the best trade-off between accuracy and computational efficiency.

\begin{table}[t!]
\centering
\caption{Model performance across different amounts of feedback data while personalizing}
\label{tab:fd}
\begin{tabular}{|c|c|}
\hline
\textbf{Training Data}               & \textbf{F1-Score} \\ \hline
Initial dataset + 5\% feedback data  & 0.80              \\ \hline
Initial dataset + 10\% feedback data & 0.85              \\ \hline
Initial dataset + 15\% feedback data & 0.88              \\ \hline
Initial dataset + 20\% feedback data & \textbf{0.91}     \\ \hline
Initial dataset + 25\% feedback data & 0.89              \\ \hline
\end{tabular}
\end{table}

\subsection{Discussion}



The experimental results demonstrate a clear trade-off between model performance and computational efficiency in the proposed personalization framework. Training models from scratch using user-specific feedback data produced the highest overall accuracy, achieving an average F1 score of 0.88. This performance improvement highlights the benefit of fully retraining the model with user-tailored motion data, which allows it to learn the unique temporal and kinematic characteristics of individual movement patterns. However, this approach comes at a significant computational cost. Training from scratch requires a longer runtime, higher memory consumption, and repeated optimization over a large parameter space, which limits its feasibility for real-time or on-device personalization.

In contrast, the transfer learning and few-shot learning approaches demonstrated much shorter personalization runtimes, making them more suitable for deployment on resource-constrained devices such as smartwatches or mobile platforms. In both cases, the lower layers of the Transformer encoder were reused to preserve general motion representations, while only the higher layers were fine-tuned on user feedback data. This strategy significantly reduced the computational burden and time required for model adaptation while maintaining reasonable accuracy. Specifically, the few-shot learning model achieved an average F1 score of 0.71 in its baseline form and 0.76 after applying the selective data strategy, outperforming the transfer learning models in both scenarios. These findings indicate that few-shot learning provides an effective balance between accuracy and computational efficiency, allowing rapid personalization without requiring extensive retraining.

Although the fully retrained models currently yield the best accuracy, their high computational cost and longer retraining time make them less practical for continuous real-world adaptation. The results suggest that few-shot learning offers the most promising direction for scalable, low-latency personalization. With further fine-tuning and optimization such as adjusting the learning rate schedule, freezing strategies, or the number of attention heads, few-shot learning could potentially approach the accuracy of full retraining while retaining its lightweight computational profile. Future work will focus on refining the few-shot pipeline to close this performance gap and to enable dynamic, on-device adaptation that operates efficiently under real-world constraints.

%% file: sections/conclusion.tex
\section{Conclusion} \label{sec:conclusion}

This study proposed a selective data-driven framework for personalization for time series data (fall detection) that integrates clustering with contrastive learning and gradient-based selection. The selective data strategy enables efficient use of limited feedback, allowing the model to focus on the most representative motion patterns while reducing redundancy and computational cost. Experimental evaluation demonstrated that incorporating selectively chosen feedback samples improves personalization performance across multiple retraining approaches, leading to more reliable and user-adaptive detection.

Overall, the proposed approach provides a scalable and efficient solution for real-time, on-device fall detection.  Building on these findings, future work will explore few-shot learning for continual personalization with minimal human supervision.